\documentclass[letterpaper, 10 pt, conference]{IEEEtran}
\IEEEoverridecommandlockouts
\usepackage{cite}
\usepackage{amsmath,amssymb,amsfonts}
\usepackage{algorithmic}
\usepackage{graphicx}
\usepackage{textcomp}
\usepackage{xcolor}
\usepackage{soul}
\usepackage[export]{adjustbox}
\usepackage{subcaption}

\def\BibTeX{{\rm B\kern-.05em{\sc i\kern-.025em b}\kern-.08em
    T\kern-.1667em\lower.7ex\hbox{E}\kern-.125emX}}
\begin{document}

\title{HSFM- $\Sigma$NN: Combining a Feedforward Motion Prediction Network and Covariance Prediction\\
}


\author{Aleksey Postnikov$^{1,2}$ \and Aleksander Gamayunov$^{1,2}$ \and Gonzalo Ferrer$^{2}$%
        \thanks{\textsuperscript{1} The authors are with the Sberbank Robotics Laboratory, Moscow, Russia.
                    {\tt\small \{postnikov.a.l,gamayunov.a.r\}@sberbank.ru}.
               }
        \thanks{\textsuperscript{2}Skolkovo Institute of Science and Technology, Moscow, Russia.
                    {\tt\small g.ferrer@skoltech.ru}.
               }    
    }

\maketitle

\begin{abstract}

In this paper, we propose a new method for motion prediction:  HSFM-$\Sigma$NN. Our proposed method combines two different approaches: a feedforward network whose layers are model-based transition functions using the HSFM and a Neural Network (NN), on each of these layers, for covariance prediction.
We will compare our method with classical methods for covariance estimation showing their limitations. We will also compare with a learning-based approach, social-LSTM, showing that our method is more precise and efficient. We will evaluate our results using the \cite{ETH,UCY} datasets.

\end{abstract}


\section{Introduction} \label{sec_intro}



High accurate prediction of human trajectories in urban environments is a topic that has been actively investigated during the last years and it has a deep impact in related topics, such as, decision making, path planning, surveillance, tracking, etc.
The problem of forecasting where pedestrians will be in the near future is, however, ill-posed by nature: Human beings tend to be unpredictable on their decisions and motion is neither exempt of it.

Most modern motion prediction algorithms focus on accurate prediction of agent position errors. Nonetheless, the precision due to this inherent uncertainty is equally important, and this paper is an effort to research on this direction.


Motion prediction algorithms has been classically divided into model and learning-based.
A more relevant classification to our paper, is based on the representation of the output:

{\em First-order moments}: usually  mean is predicted, which is a single vector of state variables. On this category we would include most of the methods.
The Social Force Model (SFM) \cite{SFM} and its Headed variant (HSFM)\cite{Farina} and \cite{Mombaur}, 
Prediction for decision making \cite{Mehta},  learning-based approaches  with deep neural networks \cite{Yi}, also learning based inverse reinforcement learning models \cite{Zhang,Fernando}
.

{\em Second-order moments}: assuming a Gaussian distribution, only two moments are required to completely specify a distribution. Many current Deep Learning (DL) approaches belong to this category, such as Social-LSTM \cite{Alahi} and other DL methods \cite{Zhang_Integrating,Gupta}.

{\em Non-parametric}  Any distribution of the prediction variables is possible, for instance an occupancy 2D grid \cite{Sarmady,Rehder}.

In this paper, we propose a motion prediction network, which propagates the system states variables, i.e. each of the pedestrians positions, over several iterations up to a time horizon.
To achieve that, we combine on {\em each} transition function (or network layer): 1) a model-based prediction (HSFM) and 2) a NN to precisely predict covariance,
getting the best of both approaches: efficiency and simplicity from model-based and precision from NN.

\begin{figure}
\centerline{\includegraphics[scale=0.32]{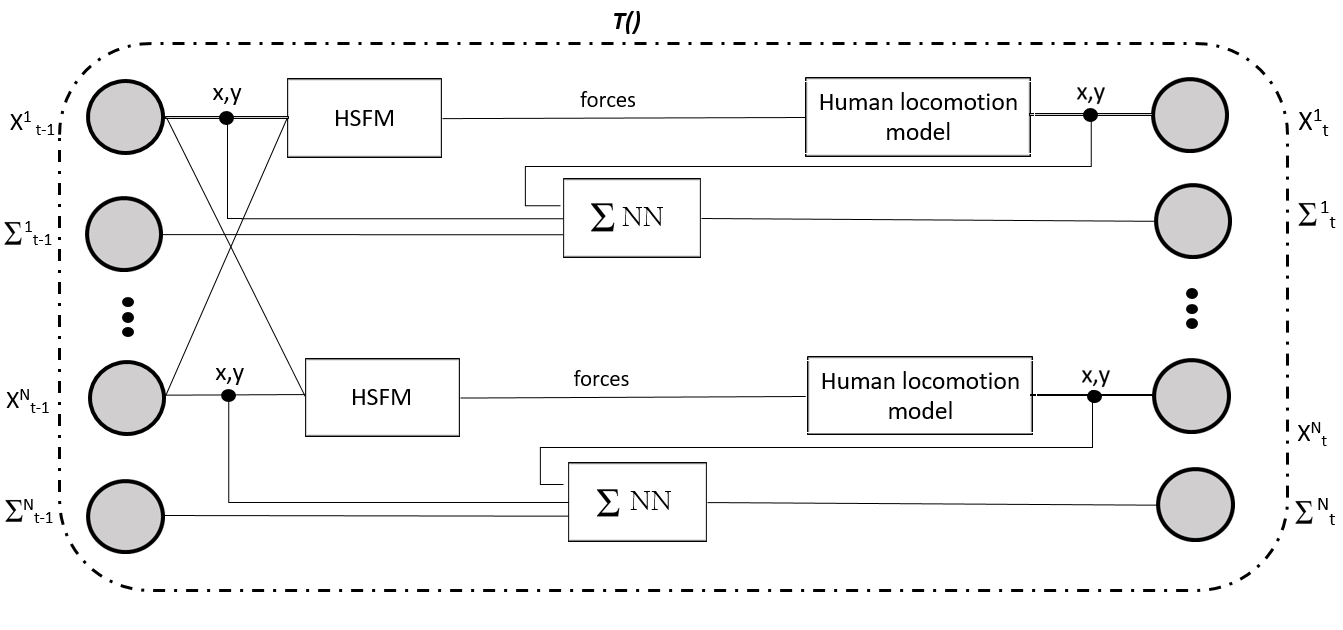}}

\caption{Diagram of the proposed method, HSFM-$\Sigma$NN. At each time-step, the HSFM generates virtual forces, which are then integrated. Our approach combines this with a NN for covariance prediction.}
\label{fig:transition}
\end{figure}
\section{Motion Prediction Network}\label{sec_pred}

In our work, we use the  transition function $T()$, shown at Fig. \ref{fig:transition}, which modifies the state variables of a pedestrian 2D pose at timestamp $t$ to $t+1$, in the following way:

\begin{equation}
x_{t+1}= T(x_{t}).
\label{eq_trans}
\end{equation}

A motion prediction network is defined as a number of consecutively stacked transition layers $T()$, similar to the feedforward network proposed in \cite{Mehta}, using SFM modules.
The contribution in this work is the addition of a shallow neural network at {\em each} transition block in order to predict covariances (Sec.~\ref{sec_covnn}).


\section{Uncertainty estimation}
\begin{figure}
    \centerline{\includegraphics[scale=0.4]{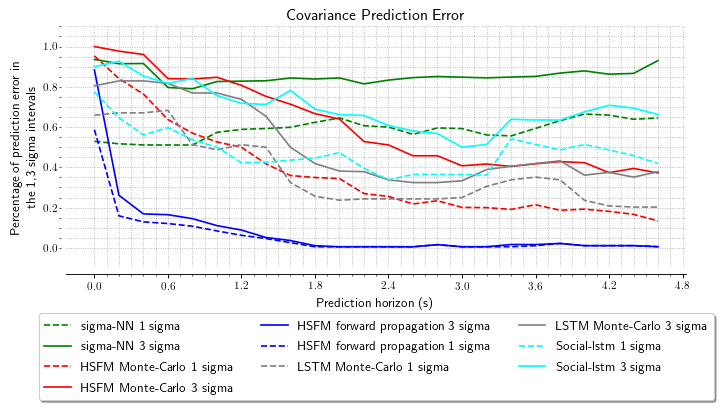}}
    \caption{Evaluation of calculated covariances for Monte-Carlo, forward propagation methods based on HSFM and social-LSTM transition functions.}
    \label{fig:3SIGMA}
\end{figure}

\subsection{Linearization and Covariance Forward-Propagation (FP)}

The transition function (\ref{eq_trans}) is a non-linear differentiable function (by construction). The simplest method for covariance estimation is using the first-order Taylor expansion:
\begin{equation}
    x_{t+1}= T(\mu_{t}) + G_{t}(x_{t}-\mu_{t}),
\end{equation}
where $\mu_{t}$ is the current state estimate and $G_{t}$ is the Jacobian of $T()$. From here, we apply {\em Covariance Propagation} of a Gaussian random variable ($x_{t} \sim \mathcal{N}( \mu_{t},\Sigma_{t})$) over a linear function:
\begin{equation}
    x_{t+1}  \sim \mathcal{N}\big(T(\mu_{t}),  G_{t} \cdot \Sigma_{t}\cdot  G_{t}^{\top}\big).
\label{eq_fp}
\end{equation}

\subsection{Monte-Carlo Covariance Estimation}\label{sec_ms}
The Monte-Carlo (MC) approach is a commonly used and powerful technique to quantify uncertainty. 

\begin{figure}
    \centerline{\includegraphics[scale=0.43]{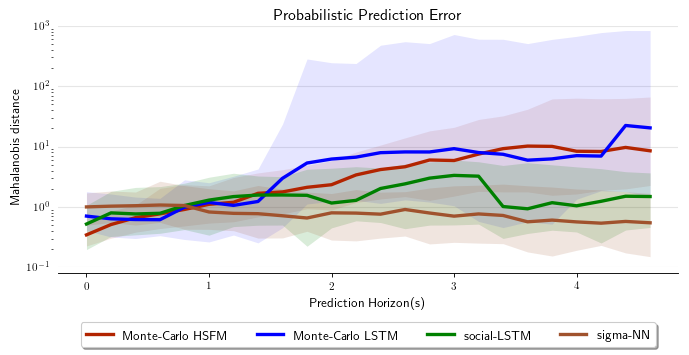}}
    \caption{Mahalanobis error distances for social-LSTM  and HSFM transition functions. In solid lines are drawn median, and colored intervals are .25 and .75 percentiles.}
    \label{fig}
\end{figure}
The procedure is straightforward: we sample from an initial distribution $x_t^i \sim p(x_t), \quad i = 1, \ldots, N$, propagate each sample $x_{t+1}^i = T(x_t^i)$ and calculate the statistics of this new set, in particular, we calculate sample mean and sample covariance.

\subsection{Neural Network Covariance Prediction}\label{sec_covnn}
In order to predict covariances, a Neural Network (NN) is trained  separately. We assume that ground truth covariances are available (see Sec.\ref{sec_exp}).
The proposed architecture is a fully connected NN, consisting of 2 hidden layers with ReLU activation function. The inputs are the stacked vectors $x_{t}$, $\Sigma_{x_{t}}$  and $x_{pred}$, as seen in Fig.~\ref{fig:transition}. 
Hidden layers input features dimensions are 100 and 50, respectively, and the final layer outputs 2 variables:  $\sigma_{x_{t+1}}^2, \sigma_{y_{t+1}}^2$.

 \section{Evaluation}
\label{sec_exp}
In this section, we present experiments on two publicly available human-trajectory datasets: ETH \cite{ETH} and UCY \cite{UCY}.

The ground truth covariance is unknown and for training purposes we approximate it as follows(which might be subject for future improvements):
\begin{equation*}
    \bar{\Sigma}_{H} = ||\tilde{x}_{1} + H \cdot v_1 - \tilde{x}_{H}||^{2}_2 \cdot  I_{2\times2}
\end{equation*}
where $v_1$ is the linear velocity at initial time, $H$ is the horizon time, $\tilde{x}_1$ and $\tilde{x}_{H}$ are obtained from the dataset(DS). This quantity is a measure on how much the future position deviates from a constant linear propagation during $H$.
Covariances are then calculated for a range of prediction horizons up to $4.8s$ and $\Delta t = 0.2s$.

Figure \ref{fig:3SIGMA} shows the results for the covariance prediction for each of the methods described above.
The graphic shows a percentage of number of times that the predicted error $||x_H - \tilde{x}_H||_{\Sigma_H}^2$, considering the estimated covariance, lies inside the $1,3\sigma$ intervals.

Then, we check the consistency of the covariance prediction by comparing with the theoretical results on 2D Gaussian variables: we should observe around 64\% of the predicted poses values being within one standard deviation interval (1$\sigma$), and  98\%  within 3$\sigma$.

\begin{table}[h!]
\centering
\begin{tabular}{||c c c||} 
 \hline
 Method & 
 \vtop{\hbox{\strut percent of predicted values}
       \hbox{\strut inside $1\sigma$ ($\Delta$ from expected) }} 
 & 
 \vtop{\hbox{\strut percent of predicted values}
       \hbox{\strut  inside $3\sigma$($\Delta$ from expected)}}\\ [0.5ex] 
 \hline\hline
 LSTM               & 47.60 (-16.39)     & 69.74 (-28.25) \\ 
 LSTM MC            & 37.16 (-26.83)     & 51.33 (-46.66) \\
 \hline\hline
 HSFM MC             & 37.11 (-26.88)    & 60.98 (-37.01) \\
 HSFM FP             & 6.06 (-57.93)   & 8.60 (-89.39) \\
 HSFM-$\Sigma$NN            & \textbf{58.79 (-5.20)}    & \textbf{85.45(-12.54)}  \\ [1ex] 
 \hline
\end{tabular}
\caption{Comparison of calculated covariances}
\label{table:1}
\end{table}

The forward propagation (FP) method collapses and provides poor results (Fig.~\ref{fig:3SIGMA}) due to vanishing gradients over multiple FPs. This is a valuable negative result we report in this paper. Stacking several layers on a prediction network makes the FP approach unusable for covariance estimation.

The MC approach is neither providing good results: for short time horizons the predicted covariance is consistent, however for larger horizons, we observe a degradation on both 1 and 3-$\sigma$, clearly underestimating the true covariance. The same result is obtained for social-LSTM.
On the other hand, our proposed method, HSFM-$\Sigma$NN achieves consistent results for any time horizon, both on 1 and 3-$\sigma$ intervals, which support the initial hypothesis of assuming Gaussian rvs and it justifies the ground truth covariance approximation.

In Fig.~\ref{fig} is depicted the Mahalanobis distance of the predicted error. In this case, we observe how both social-LSTM and our method (HSFM-$\Sigma$NN) perform well and the probabilistically weighted error norm is preserved. An unexpected drop in Mahalanobis error and covariance prediction error after 3s of forecasting for Social LSTM caused by an increase in predicted covariance. MC increases the error with the time horizon.

\section{Conclusions}
We have proposed a method, HSFM-$\Sigma$NN, for trajectory prediction based on a motion prediction network and we have added a covariance prediction NN for each of the transition modules used. We have evaluated that the most precise estimation of covariances is by NN prediction: Linear covariance propagation collapses by vanishing gradients, MC estimation does not capture the error correctly and other learning approaches, such as social-LSTM, are accurate in MH distances but become overconfident on their covariance prediction over longer horizons.


\end{document}